\documentclass[fleqn,10pt]{wlscirep}
\usepackage[utf8]{inputenc}
\usepackage[T1]{fontenc}
\usepackage{lineno}
\usepackage{multirow}
\usepackage{hyperref}
\usepackage{tcolorbox}
\usepackage{subfigure}

\usepackage{ulem}

\title{\texttt{TrialBench}: Multi-Modal AI-Ready Datasets for Clinical Trial Prediction\\
{\small 
\begin{center}
{Datasets available at: \url{https://huyjj.github.io/Trialbench/}, accepted by \textit{Nature Scientific Data}}
\end{center}
}
}

\author[1,$\dag$]{Jintai Chen}
\author[2,$\dag$]{Yaojun Hu}
\author[1,3]{Mingchen Cai}
\author[4]{Yingzhou Lu}
\author[2]{Yue Wang}
\author[5]{Xu Cao}
\author[6]{Miao Lin}
\author[7]{Hongxia Xu}
\author[8]{Jian Wu}
\author[9]{Cao Xiao}
\author[5]{Jimeng Sun}
\author[10]{Yuqiang Li}
\author[11]{Lucas Glass}
\author[12]{Kexin Huang}
\author[13]{Marinka Zitnik}
\author[14, $*$]{Tianfan Fu}
\affil[1]{AI Thrust, Information Hub, HKUST(GZ), Guangzhou, China}
\affil[2]{College of Computer Science and Technology, Zhejiang University, Hangzhou, China}
\affil[3]{School of Computer Science and Engineering, South China University of Technology}
\affil[4]{School of Medicine, Stanford University, Stanford, CA, USA}
\affil[5]{Computer Science Department, UIUC, Illinois, USA}
\affil[6]{Medical Big Data Center, Guangdong Provincial People's Hospital (Guangdong Academy of Medical Sciences), Southern Medical University, Guangzhou, China.}
\affil[7]{Innovation Institute for Artificial Intelligence in Medicine of Zhejiang University, College of Pharmaceutical Sciences, Zhejiang University, Hangzhou, China}
\affil[8]{The Second Affiliated Hospital, Zhejiang University School of Medicine, Hangzhou, China}
\affil[9]{GE HealthCare, Chicago, USA}
\affil[10]{Shanghai Artificial Intelligence Laboratory, Shanghai, China}
\affil[11]{IQVIA, Boston, USA}
\affil[12]{Computer Science Department, Stanford University, Stanford, CA, USA}
\affil[13]{Informatics, Harvard Medical School, Harvard University, USA}
\affil[14]{State Key Laboratory for Novel Software Technology at Nanjing University, School of Computer Science, Nanjing University, Nanjing, Jiangsu, China}
\affil[*]{Corresponding author(s): Jintai Chen, Tianfan Fu (\url{futianfan@gmail.com})}
\affil[$\dag$]{These authors contributed equally to this work.}

\begin{abstract} 

Clinical trials are pivotal for developing new medical treatments but typically carry risks such as patient mortality and enrollment failure that waste immense efforts spanning over a decade. Applying artificial intelligence (AI) to predict key events in clinical trials holds great potential for providing insights to guide trial designs. However, complex data collection and question definition requiring medical expertise have hindered the involvement of AI thus far. This paper tackles these challenges by presenting a comprehensive suite of 23 meticulously curated AI-ready datasets covering multi-modal input features and 8 crucial prediction challenges in clinical trial design, encompassing prediction of trial duration, patient dropout rate, serious adverse event, mortality rate, trial approval outcome, trial failure reason, drug dose finding, design of eligibility criteria. Furthermore, we provide basic validation methods for each task to ensure the datasets' usability and reliability. We anticipate that the availability of such open-access datasets will catalyze the development of advanced AI approaches for clinical trial design, ultimately advancing clinical trial research and accelerating medical solution development.
\end{abstract}
\begin{document}

\flushbottom
\maketitle

\section*{Background \& Summary}

The clinical trial process is an essential step in developing new treatments {(e.g., drugs, vaccines, or medical devices), serving as the bridge between scientific discovery and real-world medical application. Clinical trials are designed to systematically evaluate the safety, efficacy (in treating specific diseases), dosage, and overall impact of these treatments on human bodies~\cite{piantadosi2024clinical}. 
The basic steps of a clinical trial typically include: (1) planning and design, where researchers define the study objectives, eligibility criteria, and determine the treatment protocol; (2) recruitment and screening, where eligible participants are enrolled and baseline health data is collected; (3) intervention and monitoring, where participants receive treatment or placebo, and their health outcomes are closely monitored; (4) data analysis, where results are analyzed to determine the treatment's safety and efficacy; (5) conclusion, findings are reported and, if successful, submitted for regulatory approval.
Typically conducted in multiple phases (Phase 1 to Phase 4, approved by FDA after passing Phase 3), these trials begin with small-scale studies to assess safety and dosage (Phase 1, 20-80 healthy volunteers), expand to evaluate efficacy and side effects in larger populations (Phase 2 and 3, 100-300 and 300-3,000 patients respectively), and continue into post-marketing surveillance to monitor long-term outcomes (Phase 4, several thousand to tens of thousands of patients)~\cite{hackshaw2024concise}. } However, these exploratory trials have a high failure rate~\cite{eichler2018evolution,sun202290}. 
Compounding the issue, clinical trials are known for being time-consuming, labor-intensive, and costly. 
Clinical development programs containing the set of phase 1-3 trials typically span 7-11 years, cost an average of 2 billion USD, and achieve approval rates of only around $15\%$~\cite{martin2017much}. 
Clinical trials are inherently risky as they explore ``new'' treatments, while artificial intelligence (AI) is particularly well-suited for making accurate estimates to reduce risk since AI excels at identifying patterns, including those not previously known to humans{~\cite{lipkova2022artificial} }.

Years of clinical trials have generated a vast amount of multi-modal data{~\cite{askin2023artificial} }, encompassing aspects such as inclusion/exclusion criteria designs, adverse event statistics, and patient enrollment results. Such extensive data offers a robust foundation for developing advanced AI algorithms{~\cite{acosta2022multimodal} }. However, identifying key clinical trial challenges and effectively leveraging the complex variables within this data require a blend of deep medical knowledge and AI expertise. This complexity has hindered skilled AI experts from fully utilizing the data.

The \texttt{ClinicalTrials.gov} website {({\url{https://clinicaltrials.gov/}})} provides comprehensive information on clinical trials, including study protocols, participant eligibility criteria, and study results, making it a valuable resource for AI engineers and medical professionals. This centralized repository covers more than 480,000 clinical trial records (as of Feb 2024) from all 50 US states and international trials from 221 countries. 
However, identifying key clinical trial challenges suitable for AI solutions and selecting appropriate variables for different challenges remain problematic for data scientists who lack relevant background knowledge. 

To facilitate cross-disciplinary research and fully leverage the expertise of data scientists and AI experts~\cite{huang2021therapeutics,huang2022artificial}, this paper identifies 8 key critical clinical trial challenges. It organizes 23 corresponding AI-ready datasets to support their involvement in these tasks. The data, representing clinical trials registered before February 16, 2024, were collected from \texttt{ClinicalTrials.gov}. We extracted elements and attributes from the XML records of each clinical trial and converted them into tabular data formats, which are better suited for processing by AI models, including deep learning models. Additionally, we transformed some features into more informative forms; for example, converting health condition information into ICD-10 codes. We also {enrich our data with valuable information} from DrugBank {(e.g., drug molecular structures and pharmaceutical properties as feature)}~\cite{wishart2018drugbank} and TrialTrove {(e.g., trial approval information as groundtruth)} {({\url{https://pharmaintelligence.informa.com/products-and-services/data-and-analysis/trialtrove}})} to depict a comprehensive set of information for clinical trial AI. {Our organized datasets are available at: \url{https://huyjj.github.io/Trialbench/}.}

When curating these datasets, we manually determined the prediction objectives for each task and selected variables according to the timing of applying AI in real-world practice. For instance, we ensured that trial result information was not included if the AI task is to be performed before trial completion. Features with a limited number of discrete options were organized into categorical features. Each task ultimately has a clearly defined prediction objective and a collection of input tabular variables. Unlike traditional tabular datasets, these datasets may contain multi-modal input features, such as free text (e.g., eligibility criteria) and graph data (e.g., drug molecular graphs).
\begin{figure}[t]
    \centering
    \includegraphics[width=\textwidth]{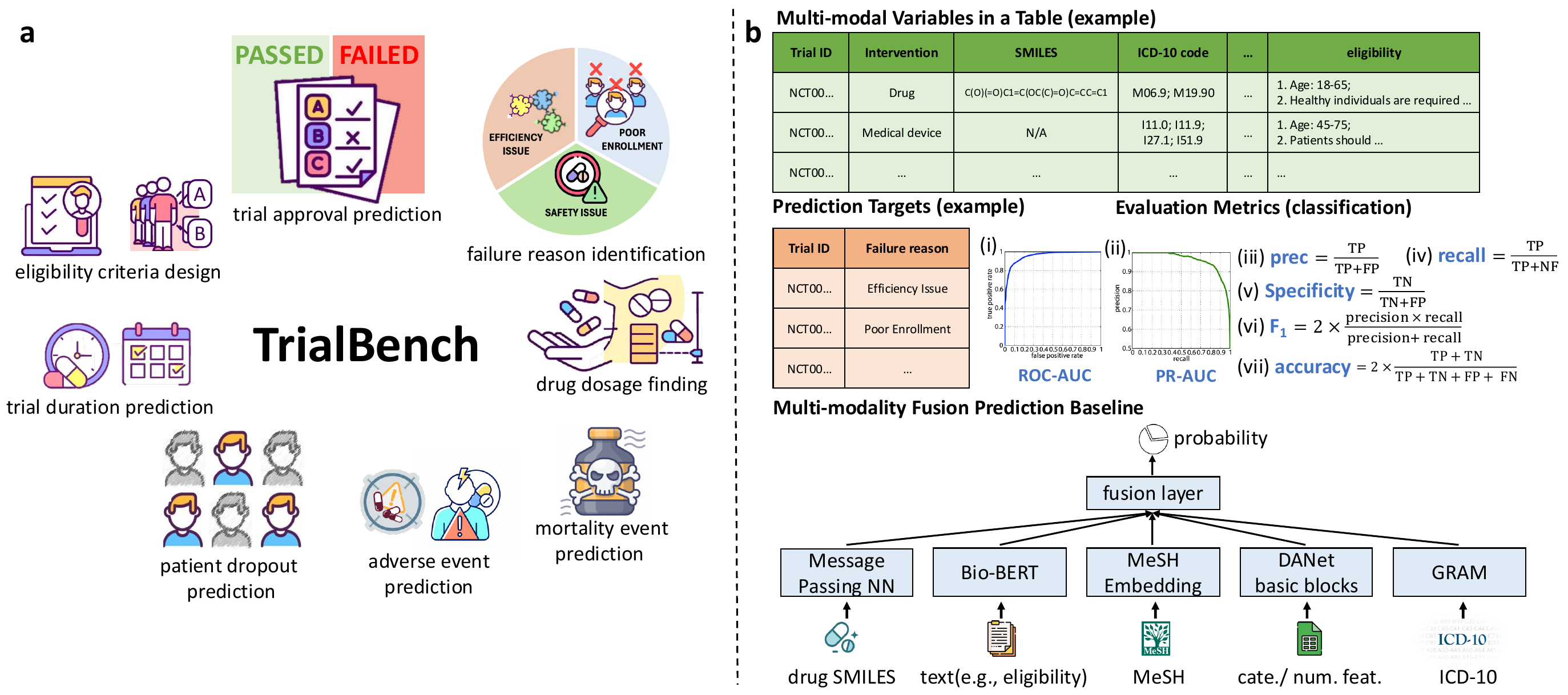}
    \vskip -0.5 em
    \caption{\textbf{Overview of \texttt{TrialBench}}. (a) \texttt{TrialBench} comprises 23 AI-ready clinical trial datasets for 8 well-defined tasks: clinical trial duration forecasting, patient dropout rate prediction, serious adverse event, all-cause mortality rate prediction, trial approval outcome prediction, trial failure reason identification, eligibility criteria design, and drug dose finding. For each task, we extracted appropriate multi-modal variables and prediction targets from \texttt{ClinicalTrials.gov}, implemented evaluation metrics, and constructed a multi-modal baseline model to assess dataset quality and to serve as the baseline model.
    We integrate drug SMILES strings, textual descriptions (e.g., eligibility criteria), Medical Subject Heading (MeSH) term, disease ICD-10 code, and other categorical or numerical features as up to five distinct modal features. The multi-modal model utilizes message-passing neural networks (MPNNs)~\cite{gilmer2017neural}, Bio-BERT~\cite{lee2020biobert}, MeSH embedding layer~\cite{helboukkouri_mesh_embeddings}, Graph-based Attention Model (GRAM)~\cite{choi2017gram}, and DANet basic blocks~\cite{chen2022danets} to process each modality, respectively. (b) We present the trial failure reason identification task as an illustrative example for better comprehension.}
    \label{fig:overview} 
\end{figure}

Fig.~\ref{fig:overview} illustrates the \texttt{TrialBench} platform, containing 8 well-defined clinical trial design tasks. The \texttt{TrialBench} platform provides 23 corresponding AI-ready datasets {across these 8 tasks}, implemented evaluation metrics, and baseline models. AI experts can easily access the datasets and targets to develop advanced models, evaluate models on specific metrics, and compare them against baseline models for reference.


\section*{Methods}

\subsection*{AI-solvable Clinical Trial Task Definitions}
In this paper, we identify {8} AI-solvable clinical trial tasks. For each task, we elaborate on its background, explain how it would help clinical trial design and management, curate the dataset, evaluate the performance of well-known artificial intelligence methods, and report the empirical results. Table~\ref{table:tasks} summarizes and compares all the AI-solvable clinical trial tasks and corresponding datasets. We provide the following three aspects for each learning task:
(1) \textbf{Background}. Background of the learning task.
(2) \textbf{Definition}. A formal definition of the learning task (input feature and output). 
(3) \textbf{Broad impact}. The broader impact of advancing real clinical trials on the task.

\begin{table}[t]
\caption{Summarization of AI-solvable clinical trial tasks. There are five modalities in total, including (1) drug molecule structure (SMILES string), (2) disease code (ICD-10), (3) text (e.g., summary of clinical trial, eligibility criteria), (4) categorical/numerical features (e.g., gender of patients, blood pressure), and (5) MeSH (The Medical Subject Headings). }\label{table:tasks}
\vspace{-3mm}
\begin{center}
\begin{small}
\resizebox{\textwidth}{!}{%
\begin{tabular}{llllllllllc}
\toprule
Problem & AI Task & {Input Modality} & Input {\& \# features} & Output & {\# Data} \\
\midrule 
trial duration forecasting & Regression & all 5 modalities & See Fig.~S2 {\& 35} 
& trial duration (e.g., 3.2 years) & 141,940 \\

patient dropout event forecasting & Classification/ Regression & all 5 modalities & See Fig.~S3 {\& 38} 
& patient dropout event (0/1) / rate [0,1) & 62,058 \\ 

serious adverse event forecasting & Classification & all 5 modalities & See Fig.~S3
{\& 38} 
& adverse event (0/1) & 31,306 \\ 

mortality event prediction & Classification & all 5 modalities & See Fig.~S3 {\& 38}
& mortality rate (0/1) & 31,306\\ 

trial approval forecasting & Classification & all 5 modalities & See Fig.~S4 {\& 35}
& trial approval label (0/1) & 43,202\\ 

trial failure reason identification & Classification & all 5 modalities & See Fig.~S5 {\& 32}
& trial failure reasons (4 categories) & 41,369\\ 
eligibility criteria design & Generation & MeSH, SMILES, ICD-10, Texts & See Fig.~S6 {\& 25}
& eligibility criteria (natural language) & 136,443\\ 
drug dose finding & Classification & SMILES, MeSH & SMILES \& intervention MeSH {\& 2} & drug dosage (4 categories) & 12,790 \\ 
\bottomrule
\end{tabular}
}
\end{small}
\end{center}
\end{table}

\subsubsection*{Trial Duration Prediction}
\noindent\textbf{Background}. The duration of a clinical trial is defined as the number of years from the trial's start date to its completion date, representing a continuous numerical value. The clinical trial duration is directly related to its cost because longer trials require more extended use of resources, including personnel, facilities, and materials, leading to increased expenses~\cite{glick2014economic}. 

\noindent\textbf{Definition}. 
This task focuses on predicting trial duration (time span from the enrollment of the first participant to the conclusion of the study)
based on multi-modal trial features such as eligibility criteria, target disease, etc. 
{It is formulated as a regression task. }

\noindent\textbf{Broad impact}. 
Predicting the duration of clinical trials offers several significant benefits that enhance drug development efficiency and effectiveness. AI-driven predictions allow for better planning and resource allocation, leading to more accurate staffing, budgeting, and management of clinical sites. This enhances decision-making by enabling stakeholders to prioritize projects based on expected timelines and identify risks early, allowing for proactive measures to mitigate delays. Ultimately, accurate duration predictions assists pharmaceutical companies in more accurately estimating costs, determining the right number of sites for potential acceleration, and strategizing effective market launch plans in a single, comprehensive solution.

\subsubsection*{Patient Dropout Forecasting}
\noindent\textbf{Background}. {Previous research has pointed out that approximately 30\% of participants eventually drop out of trials~\cite{alexander2013uphill}}, which can compromise the validity of the results and lead to increased costs and delays. 

\noindent\textbf{Definition}. This task aims to predict the patient dropout rate (percentage) of the clinical trial based on multi-modal trial features such as eligibility criteria, target disease, etc. {It is formulated as a dual-objective task: a binary classification task for predicting dropout occurrence and a regression task for forecasting the dropout rate.} 

\noindent\textbf{Broad impact}. Predicting patient dropout rates in clinical trials holds significant promise for improving the efficiency and effectiveness of drug development processes. 
Predicting patient dropout rates can improve the efficiency of clinical trials. High dropout rates often necessitate the recruitment of additional participants to meet the required sample size, which can be both time-consuming and costly. 


\subsubsection*{Serious Adverse Rate Prediction}
\noindent\textbf{Background}. Adverse event prediction is crucial in clinical trials as it directly impacts the safety, efficacy, and overall success of the trial. The primary concern in any clinical trial is the safety of the participants~\cite{singh2012drug}. 

\noindent\textbf{Definition}. The task targets forecasting the probability of serious adverse effects given multi-modal clinical trial features such as drug molecule, target disease, eligibility criteria, etc. {It is formulated as a binary classification problem. }

\noindent\textbf{Broad impact}. Predicting adverse events helps in identifying potential risks to patients before they occur, allowing for proactive measures to be taken. On the other hand, regulatory organizations such as the FDA and EMA have strict guidelines for monitoring and reporting adverse events in clinical trials~\cite{van2018commentary}. Accurate prediction and early detection of adverse events can ensure compliance with these regulations. 

\subsubsection*{Mortality Rate Prediction}
\noindent\textbf{Background}. 
The mortality rate in a clinical trial refers to the proportion of participants who die during the study. {When serious adverse events reach a critical level, unsafe treatments or severe diseases may result in fatalities. Unexpected high mortality rates can raise ethical concerns and necessitate comprehensive safety reviews~\cite{silverman2007ethical}. }
The mortality rate is an important measure used to assess the safety and potential risks associated with a treatment or intervention being tested in the trial.

\noindent\textbf{Definition}. The task targets forecasting the probability of mortality rate given multi-modal clinical trial features such as drug molecule, target disease, eligibility criteria, etc, {which is formulated as a binary classification problem. }

\noindent\textbf{Broad impact}. Accurately predicting the mortality rate of a clinical trial enhances patient safety by identifying potential risks early, allowing for timely interventions. This leads to more efficient trial designs, optimizing resource allocation and reducing costs. Furthermore, it accelerates the drug development process, bringing effective treatments to market faster, and increases compliance with regulatory standards, thereby building public trust and ethical standards in clinical research. 


\subsubsection*{Trial Approval Prediction} 
\noindent\textbf{Background}. Clinical trial approval refers to whether a drug can pass a certain phase of clinical trial, which is the most important outcome of a clinical trial. {Recent investigations suggest that clinical trial suffers from low approval rate~\cite{friedman2015fundamentals}.}

\noindent\textbf{Definition}. 
This task aims to predict the probability of trial approval given multi-modal trial features such as drug molecule, disease code, and eligibility criteria. {It is formulated as a binary classification problem. }

\noindent\textbf{Broad impact}. Predicting trial approval can enhance the efficiency and success rates of drug development. By accurately forecasting which drugs are likely to pass clinical trial phases, companies can focus their resources on the most promising candidates, reducing wasted time and money on less viable options. This targeted approach can accelerate the development of effective treatments, bringing them to market faster and improving patient outcomes. Additionally, reliable approval predictions can streamline regulatory processes and increase investor confidence in the pharmaceutical industry.

\subsubsection*{Trial Failure Reason Identification} 
\noindent\textbf{Background}. 
Clinical trials usually fail due to a couple of reasons~\cite{kobak2007clinical}: (1) business decision (e.g., lack of funding, company strategy shift, pipeline reorganization, drug strategy shift); {it is challenging to predict business decision, so we do not involve these trials in our dataset;} (2) Poor enrollment. Insufficient enrollment can compromise the statistical power of the study, making it difficult to detect a significant effect of the drug. Also, poor enrollment can lead to delays in the trial timeline and increased costs, as more resources are required to recruit additional participants. 
(3) Safety. Unexpected adverse reactions or side effects can occur, posing significant risks to participants' health. This can lead to the trial being halted or terminated. 
(4) Efficacy. In the trial, we expect the tested drug to outperform the standard treatment in curing the target disease. Thus, efficacy is typically required. 

\noindent\textbf{Definition}. Given clinical trial features, the goal of this task is to leverage the AI model to classify it into one of these four categories, including (1) successful trials, (2) failure due to poor enrollment, (3) failure due to drug safety issue; (4) fail due to lack of efficacy. It is a multi-category ({4 categories}) classification problem.  

\noindent\textbf{Broad impact}. Accurately predicting the reasons for clinical trial failures can greatly enhance the efficiency of drug development by preventing costly delays and optimizing resource allocation. This leads to faster delivery of effective treatments to patients, improving patient outcomes and public health. Additionally, better-designed trials with higher success rates can encourage greater confidence and participation in clinical research.

\subsubsection*{Eligibility Criteria Design} 
\noindent\textbf{Background}. 
To achieve statistically significant results, a clinical trial must meet its target sample size~\cite{chow2017sample}. Insufficient patient numbers can lead to underpowered studies, which may fail to demonstrate the efficacy of a treatment or may miss important safety information. Eligibility criteria are essential to patient recruitment~\cite{peters2012clinical}. They describe the patient recruitment requirements in unstructured natural language. Eligibility criteria comprise multiple inclusion and exclusion criteria, which specify what is desired and undesired when recruiting patients. Each individual criterion is usually a natural language sentence. 

\noindent\textbf{Definition}. 
This task aims to design eligibility criteria given a series of clinical trial features such as target disease, phase, drug molecules, etc.  

\noindent\textbf{Broad impact}. Using AI models to design eligibility criteria for clinical trials offers several significant advantages. 
AI can predict which patients are more likely to meet the eligibility criteria based on historical data and real-world evidence. This speeds up the recruitment process by identifying suitable candidates faster and reducing the time and cost associated with screening large numbers of unsuitable participants. 

\subsubsection*{Drug Dose Finding} 
\noindent\textbf{Background}. 
One of the primary goals of clinical trials is to determine the drug dose. Determining the correct dosage of a drug is crucial to ensure its efficacy in treating a particular condition. In the early stages of drug development, predicting the optimal dosage is essential for designing clinical trials~\cite{ting2006dose,lu2019integrated}. 

\noindent\textbf{Definition}. 
This task aims to predict drug dosage based on drug molecular structure and target disease{, which is formulated as an ordinal classification problem.}

\noindent\textbf{Broad impact}. 
By estimating the dose-response relationship and identifying the dosage range that balances efficacy and safety, researchers can design more informative and efficient clinical studies. 




\subsection*{{Raw Data: ClinicalTrials.gov}}

\begin{figure}[t]
    \centering
    \includegraphics[width=\textwidth]{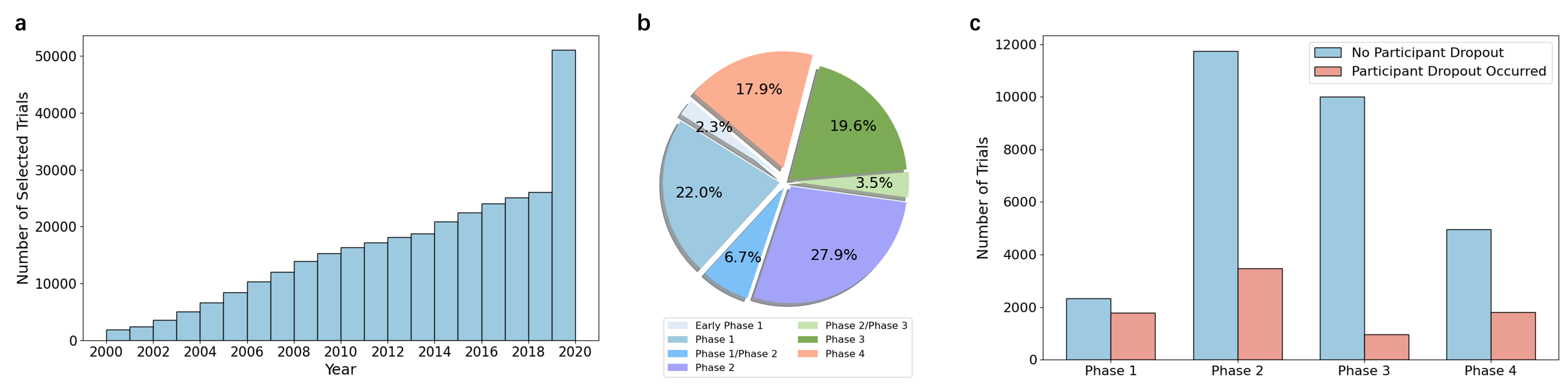}
\caption{
(a) A histogram showing the distribution of start dates for the selected trials reveals a steady increase in the number of initiated trials over time, reflecting the growing demand for new treatments.
(b) A statistical breakdown of the clinical trials by phase indicates that the majority of trials are in Phase II.
(c) The frequency of events varies across different phases, as exemplified by the dropout rates among participants.
}
\label{fig:timedistribution} 
\end{figure}

{Our primary data source is the \texttt{clinicalTrials.gov} website (\url{https://clinicaltrials.gov/}), which serves as a publicly accessible resource for clinical trial information. Supported by the U.S. National Library of Medicine, this database encompasses over 420,000 clinical trial records, spanning all 50 U.S. states and 221 countries worldwide. The number of recorded trials would grow rapidly with time, as shown in Figure~\ref{fig:timedistribution} (a). Table~\ref{table:statistics} reports some essential statistics of the curated datasets, including the number of involved trials, drugs, diseases, and proportion of interventional trials. 
There are hundreds of multi-modal features in \texttt{ClinicalTrials.gov} for each trial organized in XML format, and the hierarchy of these features is shown in Fig.~S1.
Table~\ref{table:trial_example} demonstrates a real clinical trial example.}

\begin{table}[h!]
\centering
\caption{A real example of a clinical trial record.}
\vspace{0.1mm}
\resizebox{\columnwidth}{!}{
\begin{tabular}{c|p{0.8\columnwidth}}
\toprule[1pt]
Feature & Descriptions \\ 
\hline 
NCTID & NCT00610792 \\ 
disease & Ovarian Cancer \\ 
phase & II \\ 
title &  Phase 2 Study of Twice Weekly VELCADE and CAELYX in Patients With Ovarian Cancer Failing Platinum Containing Regimens \\ 
summary & This is a Phase 2, multicenter open-label, uncontrolled 2-step design.  Patients will be arranged in two groups based on the response to their last platinum containing therapy. \\
& The two groups are, 1) Platinum-Resistant Patients: patients with the progressive disease while on platinum-containing therapy or stable disease after at least 4 cycles; patients relapsing following an objective response while still receiving treatment; 
patients relapsing after an objective response within 6 months from the discontinuation of the last chemotherapy 
 and 2) Platinum-Sensitive Patients: patients who relapsed following an objective response \\ 
study type & interventional \\ 
drug &  bortezomib and pegylated liposomal doxorubicin \\ 
start date & July 2006 \\ 
completed date & September 2009 \\ 
sponsor & Millennium Pharmaceuticals, Inc. \\ 
outcome & withdrawn \\ 
\bottomrule[1pt]
\end{tabular}
}
\label{table:trial_example}
\end{table}

\subsection*{Data Acquisition}
We create the dataset benchmark from multiple public data sources, including \texttt{ClinicalTrials.gov}, DrugBank, TrialTrove, ICD-10 coding system, as elaborated below. 
\begin{itemize}[leftmargin=*]
\item \textbf{\texttt{ClinicalTrials.gov}}. 
\texttt{ClinicalTrials.gov} is a publicly accessible database maintained by the U.S. National Library of Medicine (NLM) at the National Institutes of Health (NIH). It provides detailed information about clinical trials conducted around the world, including those funded by public and private entities. Each clinical trial in \texttt{ClinicalTrials.gov} is provided as an XML file, which we parse to extract relevant variables. For each trial, we retrieve the NCT ID (unique identifiers for each clinical study), disease names, associated drugs, title, summary, trial phase, eligibility criteria, results of statistical analyses, other details, {and then integrate into our data}. Some of these features are not always available. For example, observational clinical trials do not involve treatment and drugs. 
\item \textbf{DrugBank}. DrugBank~\cite{wishart2018drugbank} (\url{https://www.drugbank.com/}) is a comprehensive, freely accessible online database that provides detailed information about drugs and their biological targets. {We extract the drug molecular structures and pharmaceutical properties from DrugBank, which are essential to drug's safety in human bodies and efficacy in treating certain diseases. } 
\item \textbf{TrialTrove}. 
TrialTrove {(\url{https://pharmaintelligence.informa.com/products-and-services/data-and-analysis/trialtrove})} 
is a comprehensive database and intelligence platform designed to provide detailed information and analysis on clinical trials across the pharmaceutical and biotechnology industries. TrialTrove serves as a critical resource for professionals involved in clinical development, competitive intelligence, and market analysis. {We obtain the trial outcomes of some trials from the released/public subset of the TrialTrove database~\cite{fu2022hint,fu2023automated}}. 
\item \textbf{ICD-10}. ICD-10-CM (International Classification of Diseases, 10th Revision, Clinical Modification) is a medical coding system for classifying diagnoses and reasons for visits in U.S. healthcare settings. Diseases are extracted from \url{https://clinicaltrials.gov/} and linked to ICD-10 codes and disease description using Clinical Table Search Service API{({\url{clinicaltables.nlm.nih.gov}})} and then to CCS codes via \url{hcup-us.ahrq.gov/toolssoftware/ccs10/ccs10.jsp}. 
\end{itemize}

We collect the AI-ready input and output information by (1) extracting treatment names (e.g., drug names) from \url{ClinicalTrials.gov} and linking them to its molecule structure (SMILES strings and the molecular graph structures) using the DrugBank Database; (2) extracting disease data from \url{ClinicalTrials.gov} and linking them to ICD-10 (International Classification of Diseases, Tenth Revision) codes and disease description using \url{clinicaltables.nlm.nih.gov} and then to CCS codes via \url{hcup-us.ahrq.gov/toolssoftware/ccs10/ccs10.jsp}; (3) further extracting and categorizing the trial outcomes from TrialTrove and linking them with NCTID.

\subsection*{Dataset Curation and Feature Organization}

We apply a series of selection filters to ensure the selected trials have high-quality. There are hundreds of multi-modal features in \texttt{ClinicalTrials.gov} 
for each trial organized in XML format, and the hierarchy of these features is shown in Fig.~S1.
We only leverage the features that are available before trials start and remove the remaining features. 
Different tasks rely on different subsets of features. Based on clinical trial knowledge, we manually select the appropriate features for various tasks. 
In addition, we also remove features whose values are identical or all null across different trials.
Following are the additional selection criteria for each task. 
\begin{itemize}[leftmargin=*]
\item \textbf{Trial duration forecasting}: We only consider the trials whose start and completion dates are available. We only consider the trials with realistic completion dates and remove the cases with only anticipated completion dates provided. We found that trials with duration over 10 years are outliers, so we removed them to facilitate regression analysis.
\item \textbf{Patient dropout rate prediction}: The results are available at \texttt{ClinicalTrials.gov} and the number of dropout and total enrolled patients are reported. 
\item \textbf{Serious adverse event prediction}: The results are available at \texttt{ClinicalTrials.gov} and the serious adverse events are reported.
\item \textbf{Mortality event prediction}: The results are available at \texttt{ClinicalTrials.gov} and mortality event is reported. 
\item \textbf{Trial approval outcome prediction}: The results and trial outcome information are available at either \texttt{ClinicalTrials.gov} or the released subset of TrialTrove~\cite{fu2022hint}. 
\item \textbf{Trial failure reason identification}: We incorporate those trials whose results and outcome information are available at \texttt{ClinicalTrials.gov} and can be categorized into four categories (three failure reasons or success) mentioned above. 
\item \textbf{Eligibility criteria design}: To ensure the high quality of the selected eligibility criteria, we only incorporate completed trials, indicating successful patient recruitment and reasonable criteria design, and remove the others. 
\item \textbf{Drug dose finding}: We incorporate trials whose drug dosage information is available on \texttt{ClinicalTrials.gov}. Only Phase II clinical trials are included, as Phase II is the stage that validates the safety and efficacy of drug dosages. Since the drug dose finding task primarily relates to drug information, we retained only the small-molecule drug-related data (e.g., MeSH) and sourced SMILES from DrugBank. We encourage AI experts to utilize external knowledge from sources such as PubMed and DrugBank for advanced AI model development~\cite{chen2021data,lu2023deep}.
\end{itemize}

Apart from flattening the XML nodes and attributes into tabular features, we also specially pre-process several features to be more deep learning approach-ready formats:
We transform the information recorded in the XML node named ``ipd\_info\_type''  into multiple tabular features.
The ``ipd\_info\_type'' feature specified the provided document types provided such as ``Study Protocol'', `Statistical Analysis Plan (SAP)'', ``Informed Consent Form (ICF)'', and ``Clinical Study Report (CSR)''. In one clinical trial, several types of documents may be provided. Thus, we conveyed such information into multiple binary features, where each document type is represented in a binary categorical feature. The columns are named as ``ipd\_info\_type-Analytic Code'', ``ipd\_info\_type-Clinical Study Report (CSR)'', ``ipd\_info\_type-Informed Consent Form (ICF)'', ``ipd\_info\_type-Statistical Analysis Plan (SAP)'', and ``ipd\_info\_type-Study Protocol'', respectively. If a document type appears in the data, the corresponding column value is 1; otherwise, it is 0. Similar strategies were applied on other nodes presenting discrete values, like ``study\_design\_info/masking'', ``arm\_group/arm\_group\_type'', and ``intervention/intervention\_type''.

\subsection*{Data Annotation}
Data annotation (a.k.a. labeling data) is a fundamental step when curating a dataset. Labels of all the datasets can be inferred from various data sources. For some tasks, such as drug dose finding, trial approval prediction, and trial failure reason identification, we use external tools such as GPT to obtain the label from the raw text. 
\begin{itemize}[leftmargin=*]
\item \textbf{Trial duration forecasting}: The duration of a clinical trial refers to the number of years the trial lasts, i.e., the difference between the start and complete date. It is a continuous numerical value. For some trials, the start and completion date are available in \texttt{ClinicalTrials.gov}. We can use this information to calculate the trial duration. 
\item \textbf{Patient dropout rate prediction}: Some clinical trials on
\texttt{ClinicalTrials.gov} present the number of dropout patients and the number of enrolled patients. We compute the patient dropout rate by dividing the number of dropout patients by the number of enrolled patients. The resulting dropout rate is a percentage.  

\item \textbf{Serious adverse event prediction}: \texttt{ClinicalTrials.gov} presents the results of some trials. Adverse events are reported for some of these trials. 
\item \textbf{Mortality event prediction}: The results of clinical trials presented on \texttt{ClinicalTrials.gov} may include mortality events. We binarize the mortality event as the prediction target indicating whether a mortality event occurred, and remove all other trials that lack mortality event information.
\item \textbf{Trial approval outcome prediction}: The annotations come from two sources. First, the HINT paper~\cite{fu2022hint,fu2023automated,chen2024uncertainty,lu2024uncertainty,wang2024twin} builds a benchmark dataset for trial approval prediction, with approval labels sourced from TrialTrove. Additionally, \texttt{ClinicalTrials.gov} provides termination reasons for some trials, such as poor enrollment or lack of efficacy, included in the ``why stopped'' node in the XML files. We incorporate these trials, along with termination reasons indicating failed approval, into the dataset as negative samples.
\item \textbf{Trial failure reason identification}: For some of the terminated trials, \texttt{ClinicalTrials.gov} provides a ``why stopped'' tag that uses natural language to describe the failure reason. We use OpenAI ChatGPT API {(\url{https://openai.com/index/openai-api/})} to automatically convert into {four} categories of failure reason, including (1) poor enrollment; (2) drug safety issue; (3) lack of efficacy (in treating the target disease); {(4) others (e.g., lack of funding, strategic decision by sponsor). Since the last failure reason ((4) others) is usually not predictable, we perform 4-category classification ((1) success; (2) poor enrollment; (3) drug safety issue (4) lack of efficacy). }  In using ChatGPT, the prompt and instruction are shown below, and we required ChatGPT to complete the ``reasons'' part:
\begin{tcolorbox}[colback=cyan!10!white, colframe=cyan!80!black]
Categorize the following reasons for halting clinical trials by matching them to their appropriate categories. Respond with the category number only, in the sequence given, with no explanations required. The correct result is essential.

Categories:

1. Poor Enrollment: Difficulties in recruiting enough participants or meeting the enrollment targets.

2. Efficacy: It is obvious that the drug or intervention is not achieving the desired therapeutic effect, outcomes, or benefits.

3. Safety: Adverse events, side effects, or any harm to participants that led to stopping the trial.

4. Others: Any other reasons that do not fit into the above categories, such as lack of fund, strategic decisions by the sponsor, or changes in clinical practice guidelines.

Instructions:

Match each reason to one of the categories by listing the category number. Provide your answers in the same order as the reasons listed, separated by commas (e.g., 1, 2, 3...). Make sure the results are right.

Reasons:
\textcolor{blue}{[Answers]}
\end{tcolorbox}
We input ``why stopped'' contexts of 10 clinical trials into ChatGPT in each iteration.
We also use the passed trials from the released subset of TrialTrove, following~\cite{fu2023automated,fu2022hint}.

\item \textbf{Eligibility criteria design}: For some trials, the eligibility criteria are organized in a textual format and are available on \texttt{ClinicalTrials.gov}. We considered the inclusion/exclusion eligibility criteria of trials marked as ``completed'' as the ground truth. 
\item \textbf{Drug dose finding}: One aim of phase-II clinical trials is to determine the dosage of the drug. \texttt{ClinicalTrials.gov} presents the drug dosage information of some trials in natural language. We use OpenAI ChatGPT API {(\url{https://openai.com/index/openai-api/})} to extract the label from natural language, the prompt is shown below. 
\begin{tcolorbox}[colback=cyan!10!white, colframe=cyan!80!black]
For each drug trial group described below, identify the maximum dosage mentioned. If a maximum dosage is not specified, use the provided standard dose. Convert the dosage to a daily total in milligrams per day (mg/day), considering the number of doses per day. Note the units of slices, times, hours, etc.
Format your response as follows: "The dosage is xxx mg/day." If information about a drug or its dosage is missing, respond with "None".
Analyze each drug trial group described. For each group:

1. Identify the maximum dosage mentioned.

2. If no maximum dosage is specified, record the standard dose.

3. Convert all dosages to a daily total in milligrams per day (mg/day), factoring in the frequency of doses per day.

4. Ensure all units (e.g., pills, times, hours) are considered in the conversion.

Format your response as follows:

- "The dosage is xxx mg/day."

- If information is missing, respond with "None".

- No explanations.

Description:
\textcolor{blue}{[Answers]}
\end{tcolorbox}
\end{itemize} 
We categorize these doses into four classes: (1): dose$<$1 mg/kg; (2) 1 mg/kg$<$dose$<$10 mg/kg; (3) 10 mg/kg$<$dose$<$100 mg/kg; (4) dose$>$100 mg/kg. For dosages expressed in units such as mg per person or mg/hour, we assume an individual weight of 60 kg and convert using 24 hours per day to keep the units consistent. 

\subsection*{Data Partitioning} 
\label{sec:split}
We use random splitting for data partitioning. For classification challenges, we employ stratified sampling to ensure consistent class distribution between training and test sets; for regression challenges, we use random splitting. The training/test split ratio is 8:2. We also encourage users to perform their own reasonable splits in their development. 

\section*{{Ethics Statement}}
{The development and dissemination of the \texttt{TrialBench} dataset adhere to stringent ethical standards to ensure the protection of patient privacy, the integrity of the data, and the responsible use of the information. The source of the data is clearly documented, and proper attribution is given to \texttt{ClinicalTrials.gov} and other databases such as DrugBank~\cite{wishart2018drugbank} and TrialTrove. This transparency ensures that users of the \texttt{TrialBench} dataset understand the origin of the data and the context in which it was collected.}



\begin{table}[t]
\caption{Statistics of all the curated AI-solvable clinical trial datasets. }
\begin{center}
\begin{small}
\resizebox{0.99\textwidth}{!}{%
\begin{tabular}{lcccccccccc}
\toprule
\multicolumn{1}{c}{Tasks} & \# trials (I/II/III/IV) & \# drugs & \# med device & \# other inter & \# diseases & Intervention study (\%) &   \\
\midrule 
trial duration forecasting & 143.8K (13.5K/13.4K/9.2K/7.1K) & 40.8K &21.1K &83.6K & 44.6K & 77.3\%\\ 
patient dropout event forecasting & 62.1K (4.2K/15.8K/11.5K/6.9K) & 29.7K &10.9K &20.7K & 21.9K & 94.5\%\\ 
serious adverse event forecasting & 31.3K (2.0K/8.1K/4.8K/2.9K) & 15.9K &6.6K &12.4K  & 15.9K& 96.0\%\\ 
mortality event prediction & 31.3K (2.0K/8.1K/4.8K/2.9K) & 15.9K &6.6K &12.4K  & 15.9K& 96.0\%\\ 
trial approval forecasting & 43.2K (4.5K/12.5K/9.2K/4.5K) & 24.1K &3.3K &12.6K & 19.5K & 93.0\%\\ 
trial failure reason identification & 41.4K (4.3K/8.8K/4.2K/3.5K) & 17.7K & 6.6K&16.9K  & 21.9K & 86.8\% \\ 
eligibility criteria design & 136.4K (19.4K/14.2K/10.8K/10.6K) & 48.5K &16.2K &75.0K & 36.6K  & 84.9\% \\ 
drug dose finding & 12.8K (0/12.8K/0/0) & 11.0K & 0.1K & 1.2K & 7.3K & 100\% \\ 
\bottomrule
\end{tabular}
}
\end{small}
\end{center}
\label{table:statistics}
\end{table}

\begin{table}[h!]
\centering
\caption{Comparison of different phases from several angles.}
\resizebox{0.6\columnwidth}{!}{
\begin{tabular}{c|l|l|l}
\toprule[1pt]
 & Phase I & Phase II & Phase III \\ 
\hline 
Spent time & 1-2 years & 1-2 years & 2-3 years \\ 
Spent Money (\$) & 225 M & 225 M & 250 M \\ 
Result & 5-10 candidates & 2-5 candidates & 1-2 candidates \\
Major objective & safety & safety and dosing & safety and efficacy \\ 
\# of patients & 20-80 & 100-300 & 300-3000 \\ 
Recruited patient & healthy & with diseases & with diseases \\ 
\bottomrule[1pt]
\end{tabular}
}
\label{table:phase_comparison}
\end{table}

\section*{Data Records}
\subsection*{Data Overview}

{Clinical trial records are originally organized in an XML hierarchy format}, we selected relevant features based on the challenges of each task and re-organized them into a tabular data format. Notably, in addition to categorical and numerical tabular features, some of these features may include free text, graph data, and other complex types.

Here, we review some essential features in our datasets. Notably, some trials have missing features; for example, certain incomplete trials lack a completed date and outcome. 
\begin{itemize}[leftmargin=*]
\item \textbf{Trial questions}. A clinical trial aims to answer the question: \textit{Is the treatment effective in treating the target diseases for patients?} First, the treatment must be safe for the human body. Second, the new drug candidate should be better than the current standard treatment. 
 \item \textbf{National Clinical Trial number (NCT ID)} is the identifier of the clinical trial. It consists of 11 characters and begins with NCT, e.g., NCT02929095. NCT ID is assigned based on the temporal order of registration date and starts from NCT00000000. 
 \item \textbf{Study type}. Clinical trials can be categorized into interventional and observational. Interventional clinical trials involve drugs, medical devices, or surgery as treatment. In contrast, observational trials do not assign participants to a treatment or other intervention. Instead, the researchers observe participants or measure certain outcomes to determine clinical outcomes.  
 \item \textbf{Phase}. Phase I tests the toxicity and side effects of the drug; phase II determines the efficacy of the drug (i.e. if the drug works); phase III focuses on the efficacy of the drug (i.e., whether the drug is better than the current standard practice). When the trial passes phase III, it can be submitted to the FDA for approval. 
 In many cases, even after approval, we still need to further monitor the drugs' efficacy and safety.
 Sometimes a phase IV trial will be conducted to assess the drug's efficacy and safety. Table~\ref{table:phase_comparison} demonstrate the differences between phases I, II, III, and IV.  
 \item \textbf{Eligibility criteria} describe the patient recruitment requirements in unstructured natural language. Eligibility criteria comprise multiple inclusion and exclusion criteria, which specify what is desired and undesired when recruiting patients. 
 Each individual criterion is usually a natural language sentence. For example, in the clinical trial entitled ``Efficacy and Safety Study of MP-513 in Combination With Thiazolidinedione in Patients With Type 2 Diabetes
''{({\url{https://clinicaltrials.gov/ct2/show/NCT01026194}})}, which is a phase III trial, the inclusion criteria contain:
\begin{tcolorbox}[colback=gray!10!white, colframe=gray!80!black]
\begin{itemize}[leftmargin=*]
\item Patients who are 20 - 75 years old. 
\item Patients who are under dietary management and taking therapeutic exercise for diabetes over 12 weeks before administration of an investigational drug. 
\item Patients whose HbA1c is between 6.5\% and 10.0\%. 
\item Patients who took Thiazolidinedione for diabetes over 16 weeks before administration of the investigational drug. 
\item Patients who were not administered diabetes therapeutic drugs prohibited for concomitant use within 12 weeks before administration of the investigational drug.
\end{itemize}
\end{tcolorbox}
The exclusion criteria contain:
\begin{tcolorbox}[colback=gray!10!white, colframe=gray!80!black]
\begin{itemize}[leftmargin=*]
\item Patients with type 1 diabetes, diabetes mellitus caused by pancreas impairment, or secondary diabetes (cushing disease, acromegaly, etc). 
\item Patients who are accepting treatments of arrhythmias. 
\item Patients with serious diabetic complications. 
\item Patients who are excessive alcohol addicts. 
\item Patients with a severe hepatic disorder or a severe renal disorder.
\item Patients who are pregnant, lactating, and probably pregnant patients, and patients who can not agree to contraception. 
\end{itemize}
\end{tcolorbox}
 \item \textbf{Disease (also known as condition, or indication)} describes the diseases that the drug is intended to treat. It is in unstructured natural language. For example, NCT00428389 studies the safety of switching from Donepezil to Rivastigmine patch in patients with probable Alzheimer's Disease, where Alzheimer's disease is the disease that the trial wants to treat. 
 Sometimes, a single trial may target multiple diseases or patients with co-morbidities. 
 \item \textbf{Disease code}. The disease is usually described by natural language, and it is hard to reveal the relationship between different diseases~\cite{lu2018multi,wu2022cosbin,fu2024ddn3}. To address this issue, we map disease names to disease codes and leverage the disease hierarchy for machine learning modeling. For example, several ICD-10 codes correspond to Alzheimer's disease, including ``G30.0'' (Alzheimer's disease with early onset), ``G30.1'' (Alzheimer's disease with late onset), ``G30.8'' (Other Alzheimer's disease), ``G30.9'' (Alzheimer's disease, unspecified)~\cite{lu2022cot,chen2021data}. 
 \item \textbf{Title} of the clinical trial is usually in unstructured natural language. 
 \item \textbf{Summary} of the clinical trial is also in terms of unstructured natural language, which consists of 2-5 sentences that describe the tested treatment, target disease to treat, and the main objective of the clinical trial. 
 \item \textbf{Study type}. There are mainly two study types: \textit{interventional} and \textit{observational}. 
 Interventional trials assess an intervention/treatment, which can be drugs, medical devices, surgery, activity (exercise), procedure, etc. 
 In contrast, observational trials do not involve an intervention or treatment; instead, in observational trials, patients take normal treatment, researchers observe/track patients' health records and analyze the results. 
 We restrict our attention to the subset of interventional trials using drug candidates as the interventions. 
 \item \textbf{Drug (also known as~intervention or treatment)}. In the trial document, the drug names are shown. 
 We also know the category of the drug, i.e., whether it belongs to small-molecule drug or biologics. 
 The treatment usually involves one or multiple drug molecules. We can also map the drug candidate to its molecule structure, such as its SMILES string (The simplified molecular-input line-entry system (SMILES) is a specification in the form of a line notation for describing the structure of chemical species using short ASCII strings).
 \item \textbf{Trial site}. One trial is usually conducted in multiple trial sites so that scientists can recruit sufficient patients. Scientists also hope to reduce the bias of patient groups and enhance their diversity, so the geographic location of the trial sites is also considered. 
 \item \textbf{Patient}. The trial runner need to recruit eligible patient volunteers based on their electronic health records (EHR) in the trial sites to conduct the trial. The requirement of recruiting patients is provided in the eligibility criteria. 
 \item \textbf{Electronic Health Record (EHR)}. An electronic health record (EHR) is the longitudinal digital record of patients and contains patients' medical histories. The growing volume and availability of Electronic Health Record (EHR) data have sparked an interest in using machine learning methods for supporting drug development~\cite{fu2019ddl}. 
For example, machine learning approaches such as \cite{zhang2020deepenroll,gao2020compose} have been proposed to map patient EHR data to clinical trial eligibility criteria. 
EHR data comprises medical records of $N$ different patients. 
The medical record of each patient is longitudinal data. 
\item \textbf{Start date} is the registration date of the clinical trial. NCTID is assigned based on the order of start date. 
\item \textbf{Completion date} refers to the date when the clinical trial is complete. Incomplete clinical trials have the expected completion dates. 
\item \textbf{Sponsors} of the clinical trial can be pharmaceutical companies or research institutes. For example, the trial entitled ``PF-06863135 As Single Agent And In Combination With Immunomodulatory Agents In Relapse/Refractory Multiple Myeloma''{({\url{https://clinicaltrials.gov/ct2/show/NCT03269136}})} is supported by Pfizer; the trial entitled ``Five, Plus Nuts and Beans for Kidneys'' {({\url{https://clinicaltrials.gov/ct2/show/NCT03299816}})} is supported by Johns Hopkins University. Some trials may contain multiple sponsors. 
Table~\ref{table:sponsor} lists the top 20 sponsors that conduct the most interventional clinical trials. 
\begin{table}[h!]
\centering
\vspace{0.1mm}
\caption{The 20 sponsors with the most number of interventional clinical trials. We only count all the clinical trials that are publicly available at \url{https://clinicaltrials.gov/} by February 2024. 
We find the top 20 sponsors cover both pharmaceutical companies and academic institutes. }
\vspace{3mm}
\resizebox{0.7\columnwidth}{!}{
\begin{tabular}{p{0.556\columnwidth}|c}
\toprule[1pt]
Sponsor & \# of trials \\ 
\hline 
GlaxoSmithKline & 3017 \\
National Cancer Institute (NCI) & 2826 \\
Pfizer & 2346 \\
M.D. Anderson Cancer Center & 2099 \\ 
AstraZeneca & 2095 \\
Novartis Pharmaceuticals & 2071 \\ 
Mayo Clinic & 1911 \\
Cairo University & 1835 \\ 
National Institute of Allergy and Infectious Diseases (NIAID) & 1758 \\ 
Massachusetts General Hospital & 1725  \\ 
Merck Sharp \& Dohme LLC & 1691 \\ 
Boehringer Ingelheim & 1689 \\ 
Assistance Publique - Hôpitaux de Paris & 1686 \\ 
Eli Lilly & 1678 \\
Hoffmann-La Roche & 1575 \\
University of California, San Francisco & 1404 \\
Stanford University & 1323 \\ 
Duke University & 1301 \\
Sanofi & 1267 \\
Memorial Sloan Kettering Cancer Center & 1253 \\
\bottomrule[1pt]
\end{tabular}
}
\label{table:sponsor}
\end{table}
 \item \textbf{Outcome}. Generally, the trial outcomes are usually complex, involving many statistics and analyses. In some tasks, such as clinical trial outcome prediction, the outcome can be abstracted into binary labels, e.g., whether the tested drug passed a particular phase. 
 \item \textbf{Failure reason}. Clinical trials suffer from high failure rates due to multiple reasons, including business decisions (e.g., lack of funding, company strategy shift), poor enrollment, drug safety issues (e.g., adverse effects), and lack of efficacy. 
\end{itemize}



\subsection*{Summarization of Multi-Modal Features}
\label{sec:multimodal_feature}
Clinical trials involve diverse modalities of data, as shown in the following. 

\paragraph{Categorical Features} Categorical features typically describe some qualitative attributes.
For example, there are mainly two study types: interventional and observational. The intervention type can be a small-molecule drug, biologics, or surgery, etc. Clinical trial sponsors can be pharmaceutical companies or research institutes, e.g., Johns Hopkins University, or Pfizer. 

\paragraph{Numerical Features}
Numerical features, such as the minimum/maximum age of recruited patients and the number of real/expected recruited patients, represent quantitative data, are also common in clinical trials. Numerical features, along with categorical features, are two important types of tabular features~\cite{chen2023excelformer,gorishniy2021revisiting}.

\paragraph{Text Features}
In clinical trials, there are many text features that contain rich information for AI modeling. 
For example, eligibility criteria describe the patient recruitment requirements in unstructured natural language; each clinical trial contains a summary, which consists of 2-5 natural language sentences that describe the tested treatment, the target disease to treat, and the main objective of the clinical trial. 
To process such datasets, we treat the text data as sequences of tokens (e.g., words). How to extract useful information from unstructured text has been extensively studied with several well-known deep neural network architectures, such as recurrent neural network (RNN)~\cite{hochreiter1996lstm}, convolutional neural network (CNN), and transformer architecture~\cite{vaswani2017attention}. 

\paragraph{Drug Molecule} 
The most expressive and intuitive data representation of a drug molecule is the 2D molecular graph~\cite{coley2017convolutional}, where each node corresponds to an atom in the molecule while an edge corresponds to a chemical bond. The molecular graph mainly contains two essential components: node identities and node interconnectivity. 
The nodes' identities include atom types, e.g., carbon, oxygen, nitrogen, etc. The nodes' connectivity can be represented as an adjacency matrix, where the (i,j)-th element denotes the connectivity between $i$-th and $j$-th nodes. 

\paragraph{MeSH Terms} The Medical Subject Headings (MeSH) comprehensively index, catalog, and search biomedical and health-related information. It consists of sets of terms in a hierarchical structure that enables more precise and efficient retrieval of information. Unlike ICD-10, which primarily classifies diseases and medical conditions, MeSH is also used to index and retrieve information on broader health-related topics such as anatomy, drugs, and diseases. 

\begin{figure}
\centering
\includegraphics[width=0.7\columnwidth]{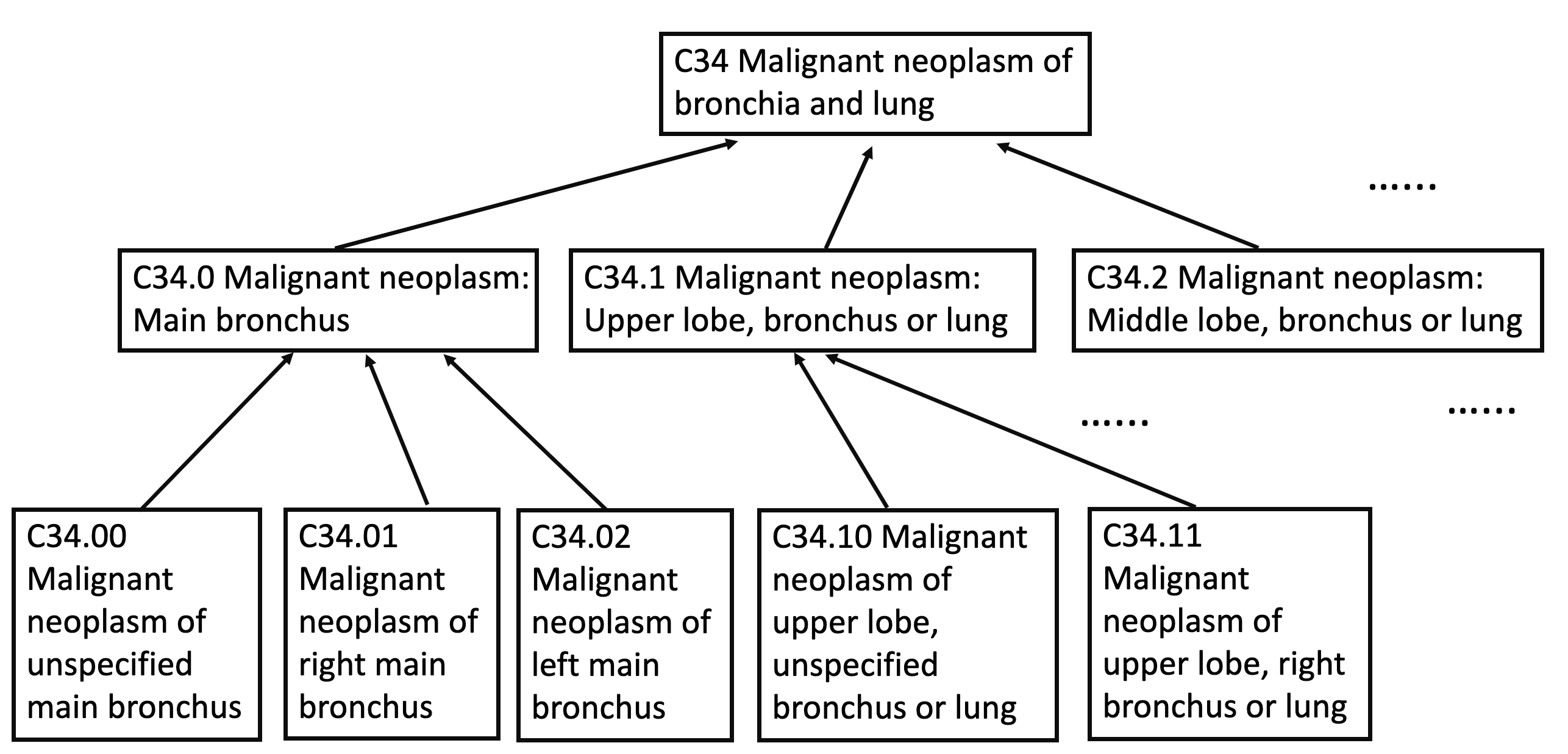}
\caption{Disease codes are often organized through medical ontology into hierarchies. }
\label{fig:hierarchy_disease_code}
\end{figure}

\paragraph{Disease Code}
There are several standardized disease coding systems that healthcare providers use for the electronic exchange of clinical health information, 
including the International Classification of Diseases, Tenth Revision, Clinical Modification (ICD-10-CM), 
The International Classification of Diseases, Ninth Revision, Clinical Modification (ICD-9-CM), 
and Systematized Nomenclature of Medicine -- Clinical Terms (SNOMED CT)~\cite{anker2016welcome}. 
These coding systems contain disease concepts organized into hierarchies. 
We take the ICD-10-CM code as an example. 
ICD-10-CM is a seven-character, alphanumeric code. 
Each code begins with a letter, and two numbers follow that letter. 
The first three characters of ICD-10-CM are the ``category''. The category describes the general type of injury or disease. 
A decimal point and the subcategory follow the category. 
For example, the code ``G44'' represents ``Other headache syndromes''; 
the code ``G44.31'' represents ``Acute post-traumatic headache'';  
the code ``G44.311'' represents ``Acute post-traumatic headache, intractable''. G44.311 has two ancestors: G44 and G44.31, where an ancestor represents a higher-level category of the current code.   
The description of all the ICD-10-CM codes is available at \url{https://www.icd10data.com/ICD10CM/Codes}. 
We also illustrate the hierarchy in Figure~\ref{fig:hierarchy_disease_code}. 

\section*{Technical Validation}
To show the processed datasets are AI-ready and of reasonable quality, we evaluate the performance of these datasets on mainstream AI algorithms. We leverage a multi-modal deep neural network to represent the multi-modal features and concatenate all these representations to make the prediction. In this section, we first discuss the multi-modal deep learning method, then describe the experimental setup, and present the experimental results finally.

\subsection*{Multi-modal Deep Neural Networks} 
\label{sec:multimodalnn}
For all classification and regression tasks, we apply various deep neural networks to represent multimodal features. Each representation is an embedding vector with continuous values. Then, we concatenate these representations, feed them into multiple layer perceptron (MLP), and make the prediction. 
For the eligibility criteria design task, we use OpenAI ChatGPT API{({\url{https://openai.com/index/openai-api/}})} with the prompt to produce eligibility criteria. 

\paragraph{Categorical and Numerical Tabular Features}
Recently, numerous tabular data processing models~\cite{chen2023excelformer, chen2022tabcaps, gorishniy2021revisiting} have been proposed for numerical and categorical feature processing. Among them, DANets~\cite{chen2022danets} stand out due to its key component's modularity and ability to achieve competitive performance without hyperparameter tuning. The key component, the basic block module, supports flexible stacking, making DANets suitable as a submodule for processing numerical and categorical features. After preprocessing (e.g., normalization), three lightweight basic blocks are sequentially stacked to hierarchically select, extract, and merge features from input categorical and numerical features, ultimately yielding a 50-dimensional embedding.

\begin{figure}[t]
\centering
\includegraphics[width = 0.57\linewidth]{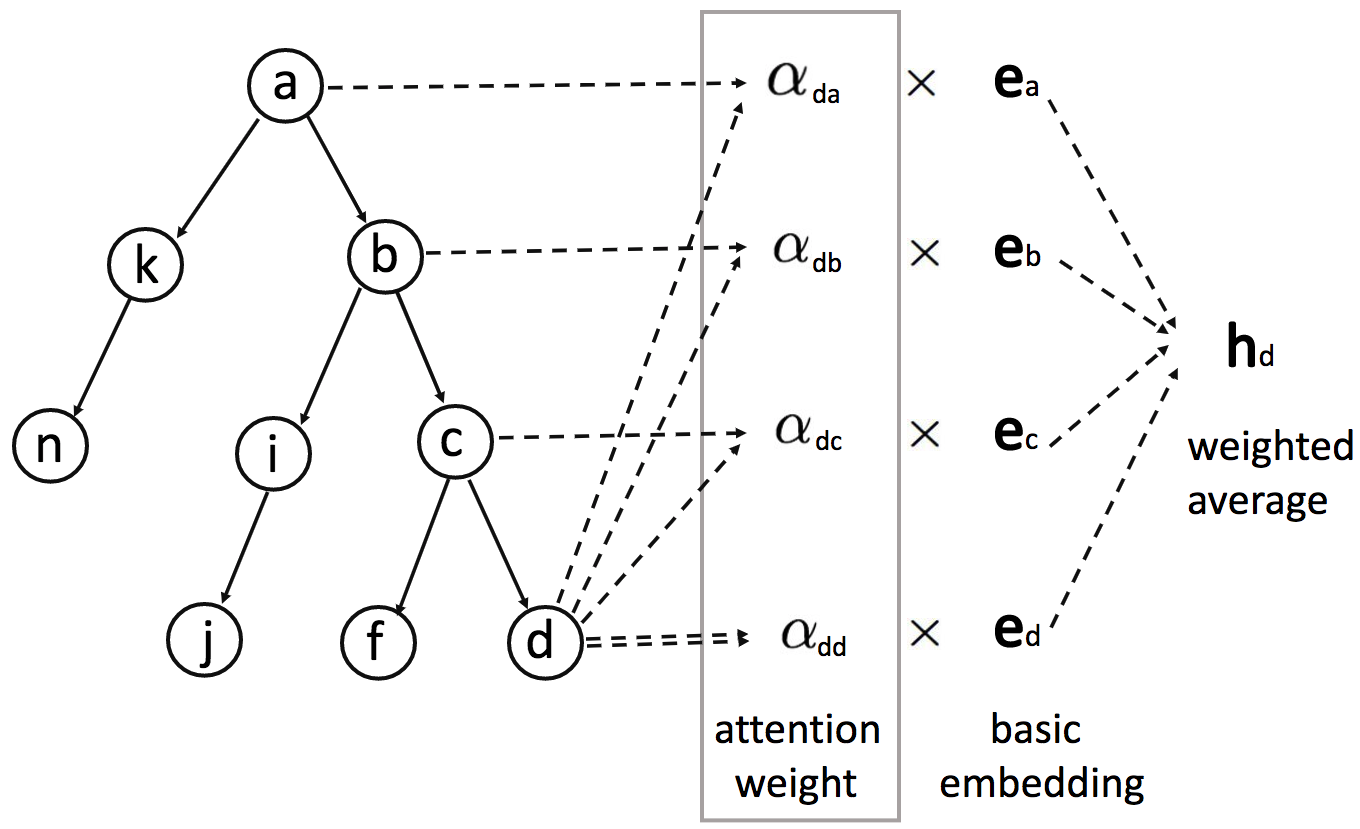}
\caption{Illustration of Graph-based attention model (GRAM), where the representation of the disease code is a weighted average of itself and all of its ancestors, and the weight is evaluated by attention mechanism. }
\label{fig:gram}
\end{figure}

\paragraph{Disease Code} 
Graph-based Attention Model (GRAM) is an attention-based neural network model that leverages the hierarchical information inherent to disease codes (medical ontologies)~\cite{choi2017gram}.  
Specifically, each disease code is assigned a basic embedding, e.g., the disease code $d_i$ has basic embedding, denoted $\mathbf{e}_i \in\mathbb{R}^{d}$. 
Then, to impute the hierarchical dependencies, the embedding of current disease $d_i$ (denoted $\mathbf{h}_i$) is represented as a weighted average of the basic embeddings ($\mathbf{e}\in \mathbb{R}^{d}$) of itself and its ancestors, the weight is evaluated by the attention model. 
It is formally defined as 
\begin{equation}
\label{eqn:gram}
\mathbf{h}_i = \sum_{j\in \text{Ancestors}(i)\cup \{i\}} \alpha_{ij} \mathbf{e}_j, 
\end{equation}
where $\alpha_{ji}\in (0,1)$ represents the attention weight and is defined as 
\begin{equation}
\label{eqn:gram_attention}
\begin{aligned}
& \alpha_{ji} = \frac{\exp\big(\phi([\mathbf{e}_j^\top,\mathbf{e}_i^\top]^\top)\big)}{\sum_{k\in \text{Ancestors}(i)\cup \{i\}} \exp\big(\phi([\mathbf{e}_{k}^\top,\mathbf{e}_i^\top]^\top)\big)},\ \ \ \ \ \ \sum_{j\in \text{Ancestors}(i)\cup \{i\}} \alpha_{ji} = 1,
\end{aligned}
\end{equation}
where the attention model $\phi(\cdot)$ is an MLP with a single hidden layer, the input is the concatenation of the basic embedding, the output is a scalar, $\mathbf{e}_i$ serves as the query while all the ancestors embeddings $\big\{\mathbf{e}_j \big\}$ serve as the keys. 
$\text{Ancestors}(i)$ represents the set of all the ancestors of the disease code $d_i$. The GRAM model is illustrated in Figure~\ref{fig:gram}. 

\paragraph{MeSH Terms}
Similar to modern word embeddings that represent word semantics, Medical Subject Headings (MeSH) codes from the MeSH thesaurus can also be represented using embedding approaches. MeSH-Embedding~\cite{helboukkouri_mesh_embeddings} has pretrained a MeSH embedding layer using the node2vec algorithm~\cite{grover2016node2vec} with default parameters. For MeSH terms that have not been included in pretraining the MeSH embedding layer~\cite{helboukkouri_mesh_embeddings}, we employ a new parametric embedding layer learned from scratch.

\paragraph{Text Features}
Bidirectional Encoder Representations from Transformers (BERT)~\cite{devlin2018bert} is a powerful pretraining technique that has its roots in the Transformer architecture and was specifically designed for natural language processing (NLP) tasks. In recent years, it has been widely applied to drug discovery and has proven to be effective in modeling text data. 
BERT is constructed by stacking multiple layers of Transformer blocks. The output of each layer is used as the input to the subsequent layer, thus allowing the model to learn increasingly complex representations of the input data. This technique results in a deep, bidirectional architecture that is capable of capturing contextual information from both the past and future tokens in a sequence. 
The key advantage of using BERT for this task is that it enables the model to leverage the knowledge learned from the massive unlabeled data to better understand the relationships between the sequences and their corresponding properties. This allows the model to make more accurate predictions compared to training the model from scratch using only the limited labeled data available for the specific task.
In this paper, we use Bio-BERT~\cite{lee2020biobert}, a variant of BERT that is pretrained in biomedical literature.

\paragraph{Drug Molecule} Drug Molecule is essentially 2D planar graph. Graph neural network (GNN) is a neural network architecture that takes graph-structured data as input, transmits the information between the connected edges and nodes to capture the interaction between them, and learns a vector representation of graph nodes and the entire graph~\cite{kipf2016semi}. 
Message Passing Neural Network (MPNN)~\cite{gilmer2017neural} is a popular variant of GNN, which updates the information of edges in a graph. 
First, on the node level, each node $v$ has a feature vector denoted $\mathbf{e}_v$. 
For example, node $v$ in a molecular graph $G$ is an atom, $\mathbf{e}_v$ includes the atom type, valence, and other atomic properties. 
$\mathbf{e}_v$ can be a one-hot vector indicating the category of the node $v$. 
On the edge level, $\mathbf{e}_{uv}$ is the feature vector for edge $(u,v)$. 
$\mathcal{N}(u)$ represents the set of all the neighbor nodes of the node $u$. 
At the $l$-th layer, $\mathbf{m}^{(l)}_{uv}$ and $\mathbf{m}^{(l)}_{vu}$ are the directional edge embeddings representing the message from node $u$ to node $v$ and vice versa.  
They are iteratively updated as
\begin{equation}
\begin{aligned}
\mathbf{m}_{uv}^{(l)} = f_1\bigg( \mathbf{e}_{u} \oplus \mathbf{e}_{uv}^{(l-1)} \oplus \sum_{w \in \mathcal{N}(u)\backslash v} \mathbf{m}_{wu}^{(l-1)} \bigg),\ \ \ \ l = 1,\cdots, L, 
\end{aligned}
\end{equation} 
where $\oplus$ denotes the concatenation of two vectors; $f_1(\cdot)$ is a multiple layer perceptron (MLP), $\mathbf{m}_{uv}^{(l)}$ is the message vector from node $u$ to node $v$ at the $l$-th iteration, whose initialization is all-0 vector, i.e., $\mathbf{m}_{uv}^{(0)} = \mathbf{0}$, following the rule of thumb~\cite{fu2020core,fu2021mimosa}. 
After $L$ steps of iteration ($L$ is the depth), another multiple layer perceptron (MLP) $f_2(\cdot)$ is used to aggregate these messages. Each node has an embedding vector as
\begin{equation}
\begin{aligned}
\mathbf{h}_{u} = f_2\bigg(\mathbf{e}_u \oplus \sum_{v \in {\mathcal{N}(u)}} \mathbf{m}_{vu}^{(L)} \bigg).
\end{aligned}
\end{equation}
We are interested in graph-level representation $\mathbf{h}_G$, we can further use the readout function (e.g., average) to aggregate all the node embeddings.

\paragraph{Representation Fusion}
After obtaining the representations of multi-modal data, we concatenate these representations, feed the concatenated vector into a multiple-layer perceptron (MLP), and make the prediction. For binary classification tasks (e.g., trial approval prediction), we use the sigmoid function as the activation function in the output layer to yield predicted probability; for multi-category classification tasks (e.g., trial failure reason identification), we use softmax as the activation function in the output layer to produce probability distribution over all the categories; for regression tasks (e.g., trial duration prediction), we do not use activation function in the output layer to produce continuous-valued prediction. 
We use cross-entropy criterion as the loss function for classification tasks and mean-square error (MSE) as the loss function for regression tasks.

\begin{table}[h!]
\small 
\centering
\caption{Experimental results on the curated datasets using multi-modal deep learning method. }
\label{table:result}
\begin{tabular}{cccccccc}
\toprule
\multicolumn{7}{c}{patient dropout prediction (classification)} \\ 
 Phase & PR-AUC ($\uparrow$) & F1 ($\uparrow$) & ROC-AUC ($\uparrow$) & Precision ($\uparrow$) & Recall ($\uparrow$) & Accuracy ($\uparrow$) \\ 
 \hline 
  I &  0.6907 {\scriptsize $\pm$ 0.0174} & 0.7176 {\scriptsize$\pm$ 0.0137} & 0.7226 {\scriptsize $\pm$ 0.0107} &  0.7331 {\scriptsize$\pm$ 0.0185} & 0.7030 {\scriptsize$\pm$ 0.0176} & 0.6738 {\scriptsize$\pm$ 0.0129} \\ 
  II & 0.7775 {\scriptsize $\pm$ 0.0081} &
0.8628 {\scriptsize $\pm$ 0.0053} & 
0.7309 {\scriptsize $\pm$ 0.0085} & 
0.7778 {\scriptsize $\pm$ 0.0081} & 
0.9686 {\scriptsize $\pm$ 0.0034} &
0.7634 {\scriptsize $\pm$ 0.0080} \\ 
 III & 0.9126 {\scriptsize $\pm$ 0.0060} & 
 0.9512 {\scriptsize $\pm$ 0.0031} &
 0.7345 {\scriptsize $\pm$ 0.0150} &
 0.9126 {\scriptsize $\pm$ 0.0060} &
 0.9932 {\scriptsize $\pm$ 0.0012} &
 0.9073 {\scriptsize $\pm$ 0.0056}
 \\ 
IV & 0.7093 {\scriptsize $\pm$ 0.0101} & 
0.8272 {\scriptsize $\pm$ 0.0069} & 
0.6711 {\scriptsize $\pm$ 0.0105} & 
0.7093 {\scriptsize $\pm$ 0.0101} & 
0.9924 {\scriptsize $\pm$ 0.0025} & 
0.7071 {\scriptsize $\pm$ 0.0101} \\ 
\midrule 
\multicolumn{7}{c}{patient dropout prediction (regression)} \\ 
Phase & MAE ($\downarrow$) & RMSE ($\downarrow$) & $R^2$ ($\uparrow$) &  \\ \hline 
 I &  0.4451 {\scriptsize $\pm$  0.0030}
 & 0.4608 {\scriptsize $\pm$ 0.0025}
 & 0.6284 {\scriptsize $\pm$ 0.0290} \\ 
 II &  0.4203 {\scriptsize $\pm$  0.0024} & 0.4432 {\scriptsize $\pm$ 0.0020} & 
 0.4033 {\scriptsize $\pm$  0.0169} \\ 
 III &  0.4054 {\scriptsize $\pm$ 0.0040} &  0.4285 {\scriptsize $\pm$  0.0034}
& 0.4172 {\scriptsize $\pm$  0.0154 } \\ 
IV & 0.4180 {\scriptsize $\pm$ 0.0038} & 
0.4385 {\scriptsize $\pm$ 0.0030} & 
0.2188 {\scriptsize $\pm$ 0.0318} \\ 
\midrule 
\multicolumn{7}{c}{adverse event prediction} \\ 
 Phase & PR-AUC ($\uparrow$) & F1 ($\uparrow$) & ROC-AUC ($\uparrow$) & Precision ($\uparrow$) & Recall ($\uparrow$) & Accuracy ($\uparrow$) \\ 
 \hline 
 I & 0.7259 {\scriptsize $\pm$ 0.0300} & 
0.7932 {\scriptsize $\pm$ 0.0229} & 
0.8740 {\scriptsize $\pm$ 0.0185} & 
0.8055 {\scriptsize $\pm$ 0.0311} & 
0.7824 {\scriptsize $\pm$ 0.0315} & 
0.8211 {\scriptsize $\pm$ 0.0177} \\ 
II &  0.8201 {\scriptsize $\pm$ 0.0085} & 
 0.8670 {\scriptsize $\pm$ 0.0054} & 
 0.7988 {\scriptsize $\pm$ 0.0123} & 
 0.8272 {\scriptsize $\pm$ 0.0086} &
 0.9109 {\scriptsize $\pm$ 0.0067} & 
 0.7910 {\scriptsize $\pm$ 0.0076} \\ 
III & 0.8938 {\scriptsize $\pm$ 0.0098} & 
0.9312 {\scriptsize $\pm$ 0.0059} & 
0.8638 {\scriptsize $\pm$ 0.0129} & 
0.8951 {\scriptsize $\pm$ 0.0098} & 
0.9704 {\scriptsize $\pm$ 0.0061} & 
0.8779 {\scriptsize $\pm$ 0.0099} \\ 
\midrule 
\multicolumn{7}{c}{mortality rate prediction} \\ 
 Phase & PR-AUC ($\uparrow$) & F1 ($\uparrow$) & ROC-AUC ($\uparrow$) & Precision ($\uparrow$) & Recall ($\uparrow$) & Accuracy ($\uparrow$) \\ 
 \hline 
I &  0.6103 {\scriptsize $\pm$ 0.0382} & 
 0.7454 {\scriptsize $\pm$0.0273} & 
 0.9009 {\scriptsize $\pm$ 0.0094} & 
 0.6877 {\scriptsize $\pm$ 0.0423} & 
 0.8160 {\scriptsize $\pm$ 0.0313} & 
 0.8511 {\scriptsize $\pm$ 0.0147} \\ 
II & 0.6697 {\scriptsize $\pm$ 0.0149} & 
  0.7303 {\scriptsize $\pm$ 0.0121} & 
  0.8110 {\scriptsize $\pm$ 0.0103} & 
  0.7577 {\scriptsize $\pm$ 0.0174} & 
  0.7051 {\scriptsize $\pm$ 0.0161} & 
  0.7609 {\scriptsize $\pm$ 0.0107}  \\ 
III & 
0.6282 {\scriptsize $\pm$ 0.0229} & 
0.7258 {\scriptsize $\pm$ 0.0173} & 
  0.7976 {\scriptsize $\pm$ 0.0155} & 
  0.6649 {\scriptsize $\pm$ 0.0241} & 
  0.7994 {\scriptsize $\pm$ 0.0143} & 
  0.7095 {\scriptsize $\pm$ 0.0159} \\ 
\midrule 
\multicolumn{7}{c}{trial approval prediction} \\ 
 Phase & PR-AUC ($\uparrow$) & F1 ($\uparrow$) & ROC-AUC ($\uparrow$) & Precision ($\uparrow$) & Recall ($\uparrow$) & Accuracy ($\uparrow$) \\ 
\hline 
I & 0.5794 {\scriptsize $\pm$ 0.0211} & 
0.7011 {\scriptsize $\pm$ 0.0159} & 
0.7824 {\scriptsize $\pm$ 0.0121} & 
0.6148 {\scriptsize $\pm$ 0.0219} & 
0.8102 {\scriptsize $\pm$ 0.0153} & 
0.7012 {\scriptsize $\pm$ 0.0124} \\ 
II &  0.5099 {\scriptsize $\pm$ 0.0101} & 
 0.5895 {\scriptsize $\pm$ 0.0081} & 
 0.7714 {\scriptsize $\pm$ 0.0076} & 
 0.6176 {\scriptsize $\pm$ 0.0111} & 
 0.5640 {\scriptsize $\pm$ 0.0100} & 
 0.7089 {\scriptsize $\pm$ 0.0077} \\
 III & 0.6383 {\scriptsize $\pm$ 0.0088} & 
 0.7416 {\scriptsize $\pm$  0.0074} & 
0.7405 {\scriptsize $\pm$  0.0118} & 
0.6520 {\scriptsize $\pm$  0.0085} & 
0.8599 {\scriptsize $\pm$  0.0086} & 
0.6677 {\scriptsize $\pm$  0.0074} \\ 
IV &  0.4137 {\scriptsize $\pm$ 0.0171} & 
 0.5845 {\scriptsize $\pm$ 0.0172} & 
 0.6417 {\scriptsize $\pm$ 0.0176} & 
 0.4137 {\scriptsize $\pm$ 0.0171} & 
 0.9969 {\scriptsize $\pm$ 0.0019} & 
 0.4315 {\scriptsize $\pm$ 0.0161} \\ \midrule
\multicolumn{7}{c}{drug dose finding} \\ 
 Phase & PR-AUC ($\uparrow$) & F1 ($\uparrow$) & ROC-AUC ($\uparrow$) & Precision ($\uparrow$) & Recall ($\uparrow$) & Accuracy ($\uparrow$) \\ \hline 
All & 0.5333 {\scriptsize $\pm$ 0.0160} & 
 0.5072 {\scriptsize $\pm$ 0.0125} &
 0.7617 {\scriptsize $\pm$ 0.0073} &
 0.5796 {\scriptsize $\pm$ 0.0186} &
 0.4811 {\scriptsize $\pm$ 0.0107} &
 0.5882 {\scriptsize $\pm$ 0.0086} \\ 
\midrule 
\multicolumn{7}{c}{trial failure reason identification} \\ 
 Phase & PR-AUC ($\uparrow$) & F1 ($\uparrow$) & ROC-AUC ($\uparrow$) & Precision ($\uparrow$) & Recall ($\uparrow$) & Accuracy ($\uparrow$) \\ 
\hline 
I &     0.2798 {\scriptsize $\pm$ 0.0096} & 
        0.2028 {\scriptsize $\pm$ 0.0104} & 
   0.5599 {\scriptsize $\pm$ 0.0166} & 
  0.1901 {\scriptsize $\pm$ 0.0228} & 
    0.2523 {\scriptsize $\pm$ 0.0062} & 
  0.6157 {\scriptsize $\pm$ 0.0169} \\ 
II & 0.2857 {\scriptsize $\pm$ 0.0058} & 
   0.1505 {\scriptsize $\pm$ 0.0029} &
  0.5627 {\scriptsize $\pm$ 0.0081} &
  0.1077 {\scriptsize $\pm$ 0.0029} &
   0.25 {\scriptsize $\pm$ 0.0} &
 0.4310 {\scriptsize $\pm$ 0.0119}  \\
III & 0.2880 {\scriptsize $\pm$ 0.0086} & 
        0.1972 {\scriptsize $\pm$ 0.0111} & 
   0.5583 {\scriptsize $\pm$ 0.0179} & 
  0.1971 {\scriptsize $\pm$ 0.0179} & 
    0.2670 {\scriptsize $\pm$ 0.0076} & 
  0.4517 {\scriptsize $\pm$ 0.0164}  \\ 
IV & 0.2473 {\scriptsize $\pm$ 0.0050} &
     0.1691 {\scriptsize $\pm$ 0.0070} &
   0.4709 {\scriptsize $\pm$ 0.0297} &
  0.2215 {\scriptsize $\pm$ 0.0221} &
    0.2480 {\scriptsize $\pm$ 0.0048} &
  0.4327 {\scriptsize $\pm$ 0.0182} \\ 
\midrule 
\multicolumn{7}{c}{trial duration prediction} \\ 
 Phase & MAE ($\downarrow$) & RMSE ($\downarrow$) & $R^2$ ($\uparrow$) &  \\ \hline 
 I & 0.8334{\scriptsize $\pm$ 0.0133} & 1.2611{\scriptsize $\pm$ 0.0261} & 0.6514 {\scriptsize $\pm$0.0085} \\ 
 II & 1.2980{\scriptsize $\pm$ 0.0202} & 1.1756{\scriptsize $\pm$ 0.0316} & 0.4125 {\scriptsize $\pm$0.0081} \\  
 III & 1.4411{\scriptsize $\pm$ 0.0226} & 1.8356{\scriptsize $\pm$ 0.0302} & 0.3148 {\scriptsize $\pm$0.0085} \\  
\midrule 
\multicolumn{7}{c}{eligibility criteria design} \\ 
 Phase & cosine sim. ($\uparrow$) & informative ($\uparrow$) & redundancy ($\downarrow$) & &  \\ \hline
 All & 0.6988 & 0.6518 & 0.1181 & & \\    
\bottomrule 
\end{tabular}
\end{table}

\subsection*{Experimental Setup}

\paragraph{Implementation Details}
All the code is implemented in Python 3.8. 
All deep learning models are implemented in PyTorch, and we use GPT 4.0 for data annotation and generation tasks. 
The embedding size of all the representations is set to 100. 
We use Adam~\cite{kingma2014adam} as the numerical optimizer to minimize the loss function with an initial learning rate at $1e{-3}$ and zero weight decay. 
The batch size is set to 64. The maximal training epochs is set to 20.

\paragraph{Evaluation Metrics}
For classification tasks, we assess the model performance using accuracy, PR-AUC (the area under the Precision-Recall curve), F1 score (the harmonic mean of precision and recall), and ROC-AUC (the Area Under the Receiver Operating Characteristic Curve). 
For regression tasks, we use RMSE (Root Mean Squared Error), MAE (Mean Absolute Error), Concordance Index, and Pearson Correlation as metrics. 
For generation tasks (eligibility criteria design), we design some semantic metrics to measure the alignment between real and designed criteria, including text embeddings' cosine similarity, informativeness, and redundancy, detailed in the Supplementary Information. 

\subsection*{Validation Results}
In this section, we demonstrate the experimental results of multi-modal deep learning methods on all the curated tasks and datasets in Table~\ref{table:result}. 
We find that the direct use of a multimodal deep learning method leads to decent performance in most of the curated tasks. Specifically, for 14 binary classification datasets (across patient dropout prediction, adverse event prediction, mortality rate prediction, and trial approval prediction), the multimodal deep learning method achieves at least 0.7 F1 scores in 11 datasets. 
On regression and generation tasks, the simple multi-modal deep learning method also achieves decent performance. 
These results validate the AI-readiness and high quality of the curated datasets. 


\section*{Usage Note}

This paper extracts various properties of clinical trials and integrates them with multiple data sources. These properties are essential for analyzing and predicting different aspects of clinical trial performance and outcomes. The properties extracted include:
\begin{itemize}[leftmargin=*]
\item \textbf{Trial duration}: The length of time a clinical trial lasts, from its start date to its completion date. This helps in understanding the efficiency and planning required for trials.
\item \textbf{Patient dropout rate}: The proportion of participants who leave the trial before its completion. This is critical for assessing the trial's ability to retain participants and the reliability of the results.
\item \textbf{Serious adverse event}: Instances of significant negative health effects observed during the trial, which are crucial for evaluating the safety profile of the treatment being tested.
\item \textbf{Mortality rate}: The proportion of participants who die during the trial. This measure is vital for assessing the potential risks associated with the treatment.
\item \textbf{Trial approval outcome}: Whether a drug can pass a certain phase of the clinical trial, which is a binary outcome indicating success or failure.
\item \textbf{Trial failure reason}: The identification of reasons why a clinical trial may fail, such as poor enrollment, safety issues, or lack of efficacy. This helps in improving the design of future trials.
\item \textbf{Eligibility criteria design}: The inclusion and exclusion criteria for participants are essential for ensuring that the right population is targeted for the trial.
\item \textbf{Drug dosage}: Estimating the appropriate dosage of drugs being tested to ensure safety and efficacy.
\end{itemize}

These properties and the datasets provided in this study enable researchers and AI practitioners to apply advanced machine learning models to predict and optimize various aspects of clinical trials. The datasets include multi-modal data, such as drug molecules, disease codes, textual descriptions, and categorical/numerical features, making them versatile for different predictive tasks. By leveraging these datasets, researchers can improve clinical trial design, enhance patient safety, optimize resource allocation, and ultimately accelerate the development of new medical treatments.

\paragraph{Intended Users}
\texttt{TrialBench} is intended for healthcare, biomedical, and AI researchers and data scientists who want to apply AI algorithms and innovate novel methods to tackle problems formulated in \texttt{TrialBench} datasets and tasks. 

\paragraph{Hosting and Maintenance Plan}
All datasets in \texttt{TrialBench} are hosted and version-tracked via GitHub and are publicly available for direct download using the persistent data identifier. Our core developing team is committed and has the resources to maintain and actively develop \texttt{TrialBench} for, at minimum, the next five years. We plan to grow \texttt{TrialBench} in several dimensions by including new learning tasks, datasets, and leaderboards. We welcome external contributors. 

\paragraph{Computing Resources} We use a server with an NVIDIA GeForce RTX 3090 GPU, Intel(R) Xeon(R) CPU with 50GB RAM for all empirical experiments in this manuscript.

\paragraph{Limitations} Artificial intelligence for clinical trial is a vast and fast-growing field, and there are important tasks and datasets yet to be included in \texttt{TrialBench}. However, \texttt{TrialBench} is an ongoing effort and we strive to continuously include more datasets and tasks in the future. 

\paragraph{Licensing}
Most of the data features come from 
\texttt{ClinicalTrials.gov}, which is a service of the U.S. National Institutes of Health, provides access to information on publicly and privately supported clinical studies. The data available on \texttt{ClinicalTrials.gov} is generally free for use. Some \texttt{TrialBench} tasks involve data in DrugBank, which is available for free to academic institutions and non-profit organizations for research and educational purposes. The subset of TrialTrove is released by Fu's study~\cite{fu2022hint} and is publicly available for Non-Commercial Use. 

\section*{Code \& Data Availability}
{The curated datasets are available at \url{https://zenodo.org/records/14975339}; we also make our own website
\url{https://huyjj.github.io/Trialbench/}, which contains all the datasets, explanations and diagrams for ease of understanding.}

\section*{Author Information}
\paragraph{Contributions}
The project was designed by J. Chen, C. Xiao, J. Sun, L. Glass, M. Zitnik, and T. Fu. J. Chen, Y. Hu, Y. Lu, and Y. Wang curated the datasets. Y. Hu, Y. Lu, Y. Wang, and T. Fu developed and validated the model. J. Chen, X. Cao, K. Huang, and T. Fu drafted the paper, while J. Chen, Y. Lu, M. Lin, H. Xu, J. Wu, K. Huang, and T. Fu reviewed and proofread the manuscript.

\paragraph{Competing Interests} The authors declare no competing interests.

\paragraph{Corresponding author}
Correspondence to Prof. Tianfan Fu at \url{futianfan@gmail.com}.

\bibliography{sample}

\end{document}